\documentclass{article}

    \usepackage[final]{neurips_2025}

\usepackage[utf8]{inputenc} 
\usepackage[T1]{fontenc}    
\usepackage{hyperref}       
\usepackage{url}            
\usepackage{booktabs}       
\usepackage{amsfonts}       
\usepackage{nicefrac}       
\usepackage{microtype}      
\usepackage{xcolor}         

\newcommand{\corrtext}{Corresponding authors: entao.yang@airliquide.com, gzhang37@cityu.edu.hk.}

\usepackage{graphicx}
\usepackage{amsmath}
\usepackage{algorithm}
\usepackage{algpseudocode}

\title{Is Grokking a Computational Glass Relaxation?}

\author{
  Xiaotian Zhang$^1$ \quad 
  Yue Shang$^{1, 2}$ \quad
  Entao Yang$^{3,}$\thanks{\corrtext} \quad
  Ge Zhang$^{1,}$\footnotemark[1] \\ 
  $^1$ Department of Physics, City University of Hong Kong, Hong Kong, China\\
  $^2$ Department of Physics and Astronomy, University of Pennsylvania, Philadelphia, PA, USA\\
  $^3$ Innovation Campus Delaware, Air Liquide, Newark, DE, USA\\
}

\begin{document}

\maketitle

\begin{abstract}
Understanding neural network's (NN) generalizability remains a central question in deep learning research.
The special phenomenon of grokking, where NNs abruptly generalize long after the training performance reaches a near-perfect level, offers a unique window to investigate the underlying mechanisms of NNs' generalizability.
Here we propose an interpretation for grokking by framing it as a \textit{computational glass relaxation}: viewing NNs as a physical system where parameters are the degrees of freedom and train loss is the system energy, we find memorization process resembles a rapid cooling of liquid into non-equilibrium glassy state at low temperature and the later generalization is like a slow relaxation towards a more stable configuration. 
This mapping enables us to sample NNs' Boltzmann entropy (density of states) landscape as a function of training loss and test accuracy. 
Our experiments in transformers on arithmetic tasks suggests that there is NO entropy barrier in the memorization-to-generalization transition of grokking, challenging previous theory that defines grokking as a first-order phase transition \cite{rubin2023grokking}.
We identify a \textit{high-entropy advantage under grokking}, an extension of prior work linking entropy to generalizability but much more significant \cite{yang2025high}. 
Inspired by grokking's far-from-equilibrium nature, we develop a toy optimizer, WanD, based on Wang-Landau Molecular Dynamics, which can eliminate grokking without any constraints and find high-norm generalizing solutions. 
This provides strictly-defined counterexamples to theory attributing grokking solely to weight norm evolution towards the Goldilocks zone \cite{liu2022omnigrok} and also suggests new potential ways for optimizer design.
\end{abstract}

\section{Introduction}
Modern deep learning generally relies on heavily overparameterized neural networks (NNs), which presents surprisingly strong generalization capabilities across diverse tasks \cite{deng2009imagenet, vaswani2017attention, devlin2019bert, brown2020language}. 
Therefore, understanding the mechanism behind this generalizability is crucial for further advancement in the field. 
While model generalization performance typically improves with the training performance simultaneously, the grokking phenomenon exhibits dramatically different dynamics, providing an intriguing case for the investigation of generalization. 
Grokking was first reported in modular arithmetic tasks where researchers used a transformer model \cite{power2022grokking}. 
The model can rapidly reach nearly 100\% training accuracy but the test accuracy remains very low, suggesting that the model only memorizes the training dataset without actually understanding them. 
However, after extensive further training, the model exhibits a sudden transition to a state of high generalization where the test accuracy is (almost) at perfect level. 
This significantly delayed generalization distinguishes grokking from  typical learning processes and raises fundamental challenges towards the existing generalization theory \cite{ power2022grokking,zhang2021understanding, kumar2023grokking}. 

Previous works have shown that the loss landscape of overparameterized NNs possesses infinitely many minima; they exhibit different generalizability and are connected with each other to form a single low-loss manifold \cite{draxler2018essentially}. 
Therefore, many studies have attempted to propose geometric indicators that correlate with the generalizability of minima. 
One representative hypothesis is that wider minima generalizes better than their narrower analogs \cite{hochreiter1997flat}. 
However, it is difficult to measure or even rigorously define the width of a minimum in such a high-dimensional space. 
While some indicators like sharpness \cite{keskar2016large} correlates with model generalization empirically, they have also been questioned for the validity in measuring generalizability alone, due to the general scaling invariance in NNs \cite{dinh2017sharp}. 
Other indicators such as the Hessian matrix are too expensive to calculate in deep learning applications \cite{keskar2016large}. 
These underscore the limitations of using local geometry alone to predict generalization, which hinders the understanding of complex generalization dynamics like grokking. 

Overcoming the limitations of purely geometric descriptions, a very recent work has explored applying concepts from statistical mechanics to NNs, which conceptualizes NN as a physical system where parameters are the degrees of freedom and uses Boltzmann entropy calculated over parameter space to characterize the loss landscape \cite{yang2025high}. 
In this context, entropy measures the \textit{logarithm of the volume} of solutions in the parameter space. 
The work reveals a \textit{high-entropy advantage} for NNs' generalizability, where the high entropy solutions (the thermodynamic equilibrium state in the context of physics) tend to exhibit better generalization than solutions obtained via classical optimizers like SGD or Adam \cite{yang2025high}. 
This motivates us to employ the entropy framework to investigate the dynamics of grokking. 
We hypothesize that the transition from memorization (pre-grokking) to generalization (grokking) might involve the network navigating towards the higher-entropy regions. 
Following the reported protocol \cite{yang2025high}, we employed the Wang-Landau Molecular Dynamics (WLMD) algorithm to sample transformer neural networks' entropy $S(ln(L_{train}),A_{test})$ as a function of the logarithm of training loss $ln(L_{train})$ and the test accuracy $A_{test}$, for three different modular arithmetic tasks with varying complexity. 
We find that the grokking process can be mapped to a computational \textit{glass relaxation}: the training loss of a NN is equivalent to the internal energy of a glass, the memorization training mimics a rapid liquid cooling to form a non-equilibrium glass at low temperature, and the eventual grokking emerges as a slow relaxation process towards a more stable, higher-entropy configuration that generalizes. 

The paper is organized as follows: In Section \ref{related_work}, we review the background of generalization, grokking, Boltzmann entropy, and glass physics. 
In Section \ref{method}, we introduce in detail how we sample the entropy landscape of an NN with WLMD. 
In Section \ref{result}, we first present entropy landscapes from three different modular arithmetic tasks, where two of them exhibits grokking while the other one is unlearnable. 
The results indicate that there is no entropy barrier between the memorization and generalization states.
Therefore, grokking is NOT a first-order phase transition, which contrasts with Rubin et al.'s previous work \cite{rubin2023grokking}. 
The grokking NNs also exhibit much more significant high-entropy advantages than those studied in Ref.~\cite{yang2025high}. 
We then eliminate grokking by training with constrained weight norm \cite{liu2022omnigrok} and find that the high-entropy advantage is significantly reduced. 
These findings inspire us to develop a toy optimizer, WanD, presented in Section \ref{wand}. 
We demonstrate that WanD can find generalizing solutions in a competitive efficiency to AdamW while almost eliminating grokking, without any weight-norm constraint or output modifications. 
Finally, in Section \ref{discussion and conclusion} we further discuss related work and situate previous findings on grokking within our proposed framework of computational glass relaxation.  We also discuss directions for future research. 
Code is available at https://github.com/xtzhang28/Grokking.

Our key contributions are as follows:
\begin{itemize}

\item \textbf{Grokking is NOT a first-order phase transition.} By conceptualizing neural networks as a physical system and sampling its entropy landscape, we find there is no entropy barrier between the pre-grokking memorization state and the final generalizing state, arguing against the first-order phase transition interpretation \cite{rubin2023grokking}.
Instead, grokking has similar behavior to the formation and relaxation of glass, where the pre-grokking memorization states can be attributed to the fast convergence of train loss minimization, and resembles the rapid cooling of a liquid during glass transition. 
The subsequent grokking process can therefore be recognized as a slow, computational glass relaxation towards a more stable, generalizing configuration.
Such an analogy can describe many reported observations for grokking and inspire potential research directions for NNs, as we detailed later.
\item \textbf{The high-entropy advantage under grokking is a unique property of generalization.} We find that in the entropy landscape of the grokking NNs, there is a huge gap between the training curve and the max-entropy curve, which was previously reported as ``high-entropy advantage in generalizability'' for other NNs and datasets \cite{yang2025high}. 
The high-entropy advantage under grokking, as we found here, is much larger. 
Furthermore, we show that after eliminating grokking by controlling the weight norm, the high-entropy advantage is much less prominent but still exists. 
This confirms that the NN without grokking does not have glassy behavior, as they are much closer to the equilibrium state, supporting the correspondence between grokking and glass relaxation. 
It also reinforces the understanding that higher entropy correlates with better NNs' generalizability \cite{yang2025high}, broadening its empirical support to our distinct model and task.
\item \textbf{We design a physics-inspired toy optimizer, WanD, which finds high-norm generalizing solutions without grokking.} WanD is an optimizer based on the Wang-Landau molecular dynamics method. 
In the modular addition task, WanD has similar time efficiency and generalization performance as AdamW but can eliminate grokking without any explicit regularization (like weight norm constraints). 
Notably, WanD often finds high-norm generalizing solutions, which challenges a theory attributing grokking solely to weight decay \cite{liu2022omnigrok}. 
This also agrees with the conclusion of Kumar et al. \cite{kumar2023grokking} that regularization is not always necessary for grokking, but in a strictly-defined situation without output scaling. 
We argue that grokking is an issue caused by optimization dynamics and can be solved by just improving the optimizer. 
We believe that WanD points out a potential direction to design new optimizers which has an advantage in finding higher entropy solutions with better generalizability, and our research can encourage further work on applying well-established statistical-physical methodologies for deep learning research. 

\end{itemize}

\section{Related Work}
\label{related_work}

\textbf{Generalization} has long been linked to flat minima in the loss landscape \cite{hochreiter1997flat}. 
In an over-parameterized NN, different minima form a connected low-loss manifold \cite{draxler2018essentially, shevchenko2020landscape}. 
Some optimizers like SGD are believed to prefer flatter minima due to its inherent gradient noise \cite{xie2020diffusion}, explaining why SGD can usually yield better generalization than other optimizers \cite{xie2020diffusion}. 
Therefore, many works focus on developing flatness-based measures and observe correlations with generalizability empirically \cite{keskar2016large, dziugaite2017computing, jiang2019fantastic, foret2020sharpness}. 
However, the usage of flatness as a simple predictor of generalization has been challenged by many works, mainly because it has the dimension of weight \cite{feng2023activity}. 
Dinh et al. \cite{dinh2017sharp} changed the loss landscape by reconstructing the fitting function without affecting the generalizability, and therefore demonstrated that sharp minima can also generalize. 
Neyshabur et al. \cite{neyshabur2017exploring} and Feng et al. \cite{feng2023activity} discussed that flatness itself is not sufficient to ensure generalization, but needs to be combined with the weight norm to obtain a suitable complexity measure. 
Our results support this because the combination of flatness and weight norm actually reflects the entropy, the logarithm of the parameter space volume, of the solutions with the same generalizability.



\textbf{Grokking} is an anomalous generalization phenomenon that was first discovered by Power et al. \cite{power2022grokking} in arithmetic tasks, describing a model's sudden transition from random-guess to (nearly) perfect test accuracy long after achieving perfect training performance. 
Different mechanisms have been proposed to explain this delayed generalization, including: 
(a) shifts from poor to good representation learning \cite{kumar2023grokking, varma2023explaining, kunin2024get, liu2022towards}; 
(b) transitions from dense to sparse predictive subnetworks \cite{merrill2023tale}; 
(c) slingshot mechanism for adaptive optimizers \cite{thilak2022slingshot}; 
(d) first-order phase transition in the network's internal representation \cite{rubin2023grokking}; 
and (e) evolution of weight norms towards the Goldilocks zone \cite{liu2022omnigrok}. 
Our analysis from the entropy perspective is consistent with (a)(b)(c), but challenges (d)(e) with well-defined counterexamples. 
Detailed discussion on connections with existing work in grokking are presented in Section \ref{discussion and conclusion}. 

\textbf{Boltzmann Entropy} defines the number of microstates for a given macrostate in a thermodynamic system. 
For example, entropy can be defined by the volume in phase space with a certain energy value: $S(U_0)=k_BlnV|_{U=U_0} $, where $k_B$ is Boltzmann constant and $V$ represents the volume in phase space that correspond to the internal energy $U_0$. 
Yang et al. \cite{yang2025high} applied the concept of Boltzmann entropy to the NNs. 
This allowed them to improve upon the local perspective of sharpness and weight norm, and established a mapping relationship between generalization and the volume of parameter space. 
They found that high-entropy NN states usually generalize better than states trained with classical optimizers like SGD, which they define as the high-entropy advantage. 
This inspired us to use the same approach to study the relationship between entropy and generalization in grokking tasks.

\textbf{Glass state} is a physical state of matter, and is usually formed by rapidly cooling or compressing a liquid. 
The glass state is not in thermodynamic equilibrium. 
Therefore, at appropriate temperatures, the glass state will undergo \textit{structural relaxation}- a process where the atomic or molecular arrangement in a non-equilibrium glass gradually evolves toward a more stable, more equilibrated configuration over time. 
This relaxation process is generally regarded as a kinetic process \cite{riechers2022predicting}. 
Winter and Janssen \cite{winter2025glassy} compared deep NNs with structural glass and found many similarities, including the relaxation dynamics at low temperatures. 
As we will show in this paper, grokking is thermodynamically similar to glass relaxation, so a promising future direction is to explore whether the conclusions in \cite{winter2025glassy} holds in our setup.

\section{Methodology}
\label{method}
In this section, we will first define the modular arithmetic task, then introduce the concept of entropy into NNs, and finally present our sampling method: the Wang-Landau Molecular Dynamics algorithm. 
Code available at https://github.com/xtzhang28/Grokking.

\subsection{Problem definition}
\label{problem definition}
The training part of our experiment is same as the setup of Nanda et al. \cite{nanda2023progress}, that is, a one-layer transformer model trained in modular arithmetic tasks. 
The data consists of two parts: prompts $\langle x\rangle\langle y\rangle\langle p\rangle\ $ and answer $\langle x\circ y\rangle$. 
In the three tasks we study, the answers are determined by the following equations:

\begin{equation}
    \begin{aligned}
    &x \circ y=x+y\ mod\ p\ \mathrm{for}\ 0\leq x,y<p\\
    &x \circ y=x^2+y\ mod\ p\ \mathrm{for}\ 0\leq x,y<p\\
    &x \circ y=x^3+xy^2+y\ mod\ p\ \mathrm{for}\ 0\leq x,y<p
    \end{aligned}
    \label{modular arithmetic tasks}
\end{equation}

Here, $x$, $y$, and $p$ are all natural numbers. 
In this paper, $p$ is set to a prime number 67. 
The entire dataset will be divided into a training set and a test set with a 50\% fraction, by a fixed random seed. 
In the Appendix Figure \ref{fig:visual}, we provide a visual image of the full dataset. 
Transformer hyperparameters and more training details are also given in the Appendix \ref{additional detials}.

\subsection{Entropy in neural networks}
\label{entropy in nns}
There are many ways to define entropy in physics and information theory. 
In this work, we apply the definition of Boltzmann entropy $S=k_B\ln V$ to NNs. 
Here, $k_B$ is the Boltzmann constant, which we set to 1 for simplicity. 
In physics, $V$ is the volume of the phase space that corresponds to a particular set of observables (energy, magnetization, etc.).
For an NN, we define $V$ as the volume of the parameter space that corresponds to a particular training loss and test accuracy, and therefore $S$ is also a function of these two quantities.
To ensure that $V$ is finite, we limit the parameter space by setting different reflective bounds for the parameters of different layers.
These bounds are designed to cover the range of parameters typically reached during normal training while avoiding being excessively large.
In Sec.~\ref{advantage_without_grokking}, we avoid applying these reflective bounds to study the entropy advantage when grokking is suppressed. 
Instead, we limit our study to a hyperspherical shell of weight norm 30, a method known to effectively prevent grokking \cite{liu2022omnigrok}.
Considering grokking often occurs when the training loss is small, our results are presented in terms of its logarithm, $S(ln(L_{train}),A_{test})$.
Here, $L_{train}$ denotes the training cross-entropy loss, and $A_{test}$ represents the smoothed test accuracy.

The entropy landscape we calculate cannot reach infinite resolution. 
For computational purposes, we divide $S(ln(L_{train}),A_{test})$, which is a 2D function, into rectangular bins, where each bin corresponds to the entropy of the NN parameters within a certain range of training loss and test accuracy. 
The Wang-Landau Molecular Dynamics method allows us to calculate $S(ln(L_{train}),A_{test})$ with an arbitrary zero point. 
In other words, the absolute value of the entropy is meaningless, but the entropy difference between any two bins is meaningful.
Therefore, we can subtract a constant from the entropy in each bin so that the minima-bin entropy is always 0.

\subsection{Wang-Landau Molecular Dynamics (WLMD)}
\label{WLMD method}
In this section, we will introduce the details of WLMD algorithm and its implementation in NNs.
The key advantage WLMD is its ability to avoid becoming trapped \cite{junghans2014molecular}. 
In contrast to standard samplers, which tend to oversample the high-entropy regions, WLMD achieves uniform sampling across all regions of the entropy diagram upon convergence \cite{kim2006statistical}.
This is accomplished by maintaining a running estimate of the entropy landscape, which introduces an ‘entropy gradient’ that drives the system away from previously visited states.
The convergence of the WLMD is established in \cite{fort2015convergence}.
For a physical system made of atoms or particles of other length scales, one can use WLMD to calculate the entropy as a function of the potential energy $S(U)$. 
This is done by simulating particles with the following equation of motion \cite{junghans2014molecular}:
\begin{equation}
    \begin{aligned}
    \frac{d^2q_i}{dt^2}=-\frac{T_0}{m_i}\frac{\partial S(U,t)}{\partial U}\frac{\partial U(q_1,q_2,...,q_n)}{\partial q_i},
    \end{aligned}
    \label{partical motion}
\end{equation}
where $q_1,q_2,...,q_i$ is the coordinate of each particle, $T_0$ is the simulation temperature, $m_i$ is the mass of the particle, $t$ is the simulation time, $S(U, t)$ is an entropy function that we update over time, and $U$ is the potential energy of the system. 
Initially, $S(U,t=0)=0$. During the simulation, we repeatedly update $S(U,t)$:
\begin{equation}
    \begin{aligned}
    S(U,t+\Delta t)=S(U,t)+f_{WL}(t)\delta (U-U_0),\ 
    \mathrm{where}\  f_{WL}(t)=F_{WL}\ \mbox{min}\{t/t_{WL},t_{WL}/t\}.
    \end{aligned}
    \label{entropy update}
\end{equation}
Here the Wang-Landau factor $f_{WL}(t)$ quantifies how fast we update the entropy function.
In order to ensure the convergence of the WLMD algorithm, usually we make it increase linearly with time to a maximum value $F_{WL}$ and then decay inversely. 
$\delta$ is a Dirac function, and we approximated it with a narrow Gaussian function in the simulation. 
$U_0$ is the potential energy of the system at this moment.

For NNs, we can set the temperature $T_0$ and the mass $m_i$ to 1, which does not affect the correctness of the WLMD algorithm. 
The particle coordinates $q_1,q_2,...,q_n$ are replaced by the NN parameters $c_1,c_2,...,c_n$, and the entropy we considered is $S(x_{WL},y_{WL})$. We choose $ln(L_{train})$ as the energy in a broad sense $x_{WL}$, and $A_{test}$ as the order parameter $y_{WL}$. 
The parameter motion follows
\begin{equation}
    \begin{aligned}
    \frac{d^2c_i}{dt^2}=-\frac{\partial S(x_{WL},y_{WL},t)}{\partial c_i}=-\frac{\partial S(ln(L_{train}),A_{test},t)}{\partial ln(L_{train})}\frac{\partial ln(L_{train}(c_1,c_2,...,c_n))}{\partial c_i}\\
    -\frac{\partial Sln(L_{train}),A_{test},t)}{\partial A_{test}}\frac{\partial A_{test}(c_1,c_2,...,c_n)}{\partial c_i}.
    \end{aligned}
    \label{parameter motion}
\end{equation}

The term $\frac{\partial S}{\partial ln(L_{train})}$ and $\frac{\partial S}{\partial A_{test}}$ can be computed via numerical differentiation between adjacent bins, while $\frac{\partial ln(L_{train})}{\partial c_i}$ is the gradient of $ln(train\ loss)$ over parameters. Since $A_{test}$ is discrete, we approximate its gradient using the sigmoid function, following the approach in \cite{yang2025high}.
Further details are also provided in Appendix \ref{WLMD details}.
Reflective boundary conditions are used to implement parameter constraints for results in Sec.~\ref{no_entropy_barrier}, and for results in Section \ref{advantage_without_grokking} we restricted the parameter weight norms directly by weight rescaling (both detailed in Sec.~\ref{entropy in nns}).

The initialized velocity of the parameters follows the Maxwell–Boltzmann distribution, which is a random Gaussian distribution with mean $0$ and variance $k_BT/m=1$. 
In order to maintain constant temperature during the simulation, that is, to conserve the kinetic energy of the parameters, we add the Langevin kinetic term to the velocities:
\begin{equation}
    \begin{aligned}
    \frac{dc_i}{dt}(t+\Delta t)=\frac{dc_i}{dt}(t)+\frac{d^2c_i}{dt^2}\Delta t-c_f\frac{dc_i}{dt}(t)+\sqrt{c_f(2-c_f)k_BT}v_{rand}
    \end{aligned}
    \label{Langevin motion}
\end{equation}
Here, the definition of $\frac{d^2c_i}{dt^2}$ is given in Equation \ref{parameter motion}, $c_f$ is the friction coefficient, and $v_{rand}$ is the random velocity follows a standard Gaussian distribution. 
In the NN parameters updating process, by continuously adding entropy to the entropy landscape, $S(x_{WL},y_{WL},t)$ will eventually converge to the true entropy distribution $S(x_{WL},y_{WL})$.

Due to space limitations, more details of WLMD and the complete algorithm are described in the Appendix \ref{WLMD details}.
To the best of our knowledge, WLMD is currently the most efficient entropy sampling algorithm, as existing alternatives are limited to models with only hundreds of parameters \cite{yang2025high}. 
The computational complexity of WLMD scales linearly with the number of NN parameters, and the entropy landscape sampling can be parallelized across multiple GPUs.
This enables entropy sampling for large-scale models, such as transformers with hundreds of thousands of parameters.

\section{Grokking: Glass Relaxation Process in Neural Network}
\label{result}

\subsection{No entropy barrier in entropy landscape}
\label{no_entropy_barrier}


In thermodynamics, a phase transition is marked by a significant change in an order parameter—a macroscopic, measurable quantity like density during a liquid–gas transition. 
By analogy, the learning dynamics of NNs exhibiting grokking appears to mimic the characteristics of a phase transition:
starting with a rapid progress from underfitting to memorization (low test accuracy), the NN undergoes an abrupt qualitative shift to generalization (high test accuracy) after a prolonged training period.
Given this, we identify the test accuracy as the order parameter for characterizing grokking. 
Just as a physical order parameter changes distinctly between material states, the definitive increase in test accuracy can track the NN's transition into the generalization regime, making it a natural and compelling macroscopic indicator of this critical learning phenomenon.

\begin{figure}[htb]
  \centering
  \includegraphics[width=\linewidth]{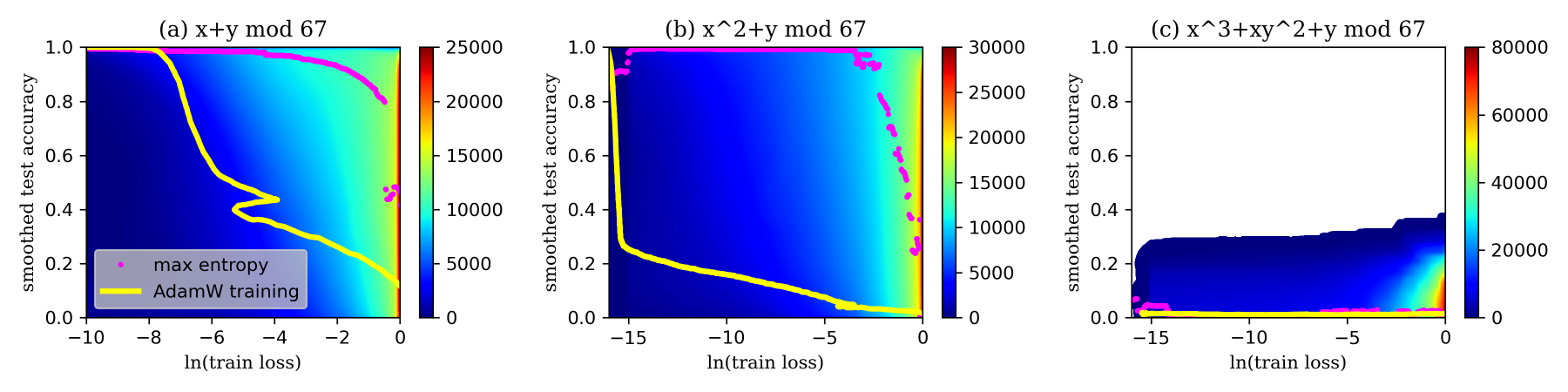}
  \caption{Entropy landscape of three different arithmetic tasks : (a) $x+y\ mod\ 67$ is easy to learn, max entropy state can generalize better than training line. (b) $x^2+y\ mod\ 67$ is hard to learn, the generalization advantage of the max entropy state over the training results under the same training loss is very significant. (c) $x^3+xy^2+y\ mod\ 67$ cannot be learned by training, max entropy state also can not generalize. }
  \label{fig:entropytasks}
\end{figure}

A previous study suggests that grokking is a first-order phase transition process from memorization to generalization \cite{rubin2023grokking}. 
They used the Langevin dynamic optimizer to train two two-layer nonlinear teacher-student models and found that the memorization and generalization states of the NN were located at two free-energy minima, thus believing that grokking is a first-order phase transition that requires crossing a free-energy barrier. 
We directly examined this statement by computing the entropy landscape as a function of training loss and test accuracy to search for such a barrier.
We examine three algorithmic tasks on a one-layer transformer model. 
The three tasks range from easy to learn, to difficult to learn, and to impossible to learn. 
The training results are also given in the Appendix Figure \ref{fig:three_tasks_training}, where grokking is observed for the first two tasks. 
Using the WLMD method, we sample the entropy landscape of training loss and test accuracy as presented in Figure \ref{fig:entropytasks}.
The three entropy landscapes are the results after 60M, 80M, and 20M WLMD update epochs respectively, where these durations were chosen as the points where the entropy map exhibited no substantial further change. 
We ran 8 WLMD processes on 4 GeForce RTX 4090 GPUs, requiring approximately 100 GPU hours per 1M epochs. 
Additional computational details are provided in Appendix \ref{training_details_and_cost}.

For the first two tasks that exhibit grokking, the training trajectory is roughly ``L'' shaped. In the vertical part of the trajectory, i.e., from the memorization state (lower left part) to the generalization state (upper left part), we observed no entropy barrier. 
Since there is neither an entropy barrier nor an energy (train loss) barrier, our results disproves the view that grokking requires crossing a free-energy barrier. 
In addition, we find that the equilibrium (maximum entropy for a given training loss) curve in the low train loss region is continuous, which also denies the first-order phase transition, as such transitions require discontinuity of the order parameter (test accuracy). 
For the third unlearnable task, the equilibrium state also fails to generalize, although some states present a higher test accuracy (about 20\%) than random selection (about 1.5\%).

We also found a huge gap between the AdamW training curve and the equilibrium state curve. 
We define this significant advantage of the equilibrium state over the training curve as the high-entropy advantage under grokking. 
This result shows that grokking is actually a kinetic process rather than a phase transition in the thermodynamic sense: The NN is far  from the equilibrium state during the initial part of the training process, and finally reaches the equilibrium state, which generalizes well, after a long time. 
This process is similar to glass formation and relaxation: When the liquid is cooled rapidly, the molecules do not have enough time to rearrange themselves into a crystal or crystal-like structure, forming a disordered glassy state (non-equilibrium state), which then gradually evolves toward a better equilibrated state through slow structural relaxation.

\subsection{Neural network without grokking does not exhibit glassy behavior}
\label{advantage_without_grokking}

The analogy between grokking and glass relaxation can be further verified by examining the entropy landscape after eliminating grokking. 
According to our conjecture, just as a slowly cooled liquid will maintain thermodynamic equilibrium and will not enter the non-equilibrium glassy state, an NN without grokking will be trained along the equilibrium curve as the training loss decreases, steadily reaching the generalization state. 
The study by Liu et al. \cite{liu2022omnigrok} suggested that grokking can be eliminated by limiting the weight norm of the NN parameters in an appropriate range. 
Following this insight, we implemented a strict weight norm constraint in our study: after each iteration step, we compute the weight norm of the neural network parameters and scale them proportionally to maintain a fixed norm of 30.
The training results, presented in Appendix Figure \ref{fig:AdamW_training_result}, show that grokking can be greatly eliminated under this condition. 
We examine three transformer models with varying widths (model dimensions).
For each entropy sampling task, we use 4 NVIDIA H100 GPUs to run 8 processes, totaling 60M epochs and consume nearly 4,000 GPU hours. 
Further computational details are available in Appendix \ref{training_details_and_cost}.

\begin{figure}[htb]
  \centering
  \includegraphics[width=\linewidth]{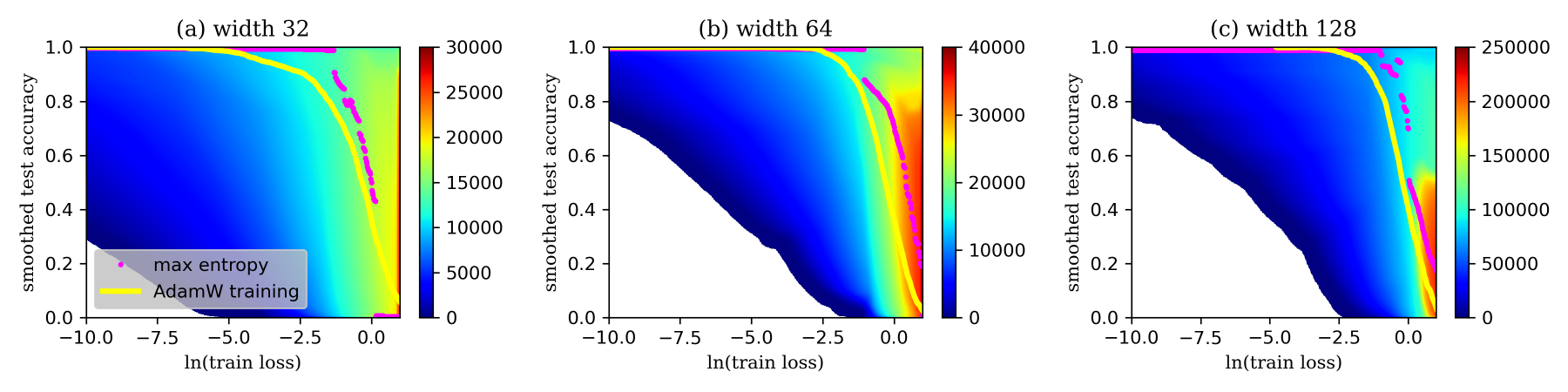}
  \caption{Entropy landscape of modular addition ($p=67$) task, sampling in three transformer models that have different widths. Here weight norm is constrained to a fixed value of 30. Compared with the grokking NNs, the high-entropy advantage is significantly weakened but still exists. }
  \label{fig:entropywidths}
\end{figure}

Figure \ref{fig:entropywidths} shows that after controlling the weight norm at 30, the AdamW training curve in the entropy landscape is very close but still slightly below the equilibrium state. 
Moreover, under this restriction the entropy landscape cannot explore severe memorization states, which is the computational glassy state in our analogy. 
This result further illustrates the correspondence between glass and grokking. 
In addition, we note that the high-entropy advantage still exists even in the network where grokking is eliminated. 
This agrees with the work of Yang et al. \cite{yang2025high} that high-entropy advantages are prevalent in various deep learning tasks and there is a robust connection between equilibrium and generalization.

\section{WanD: Eliminating grokking without regularization}
\label{wand}

The above results show that the equilibrium state at low training loss has high generalizability regardless of whether the weight norm of the network is restricted. 
Therefore, we believe that grokking is a dynamic process caused by the optimizer's inefficient exploration of the loss landscape basins. 
The WLMD method we used is good at sampling high entropy states of NNs, which inspires us to design an optimizer by imitating WLMD, that can find high entropy states and therefore should also eliminate grokking. 
Here, we design an optimizer, WanD, based on the one-dimensional Wang-Landau Molecular Dynamics. The continuous-time dynamics of WanD is as follows.
\begin{equation}
    \begin{aligned}
    &\frac{d^2\theta(t)}{dt^2}= -\frac{\partial S(ln(L_{train}(\theta(t))))}{\partial ln(L_{train}(\theta(t)))}\cdot \frac{\partial ln(L_{train}(\theta(t)))}{\partial \theta(t)}+\eta(t)-\lambda\frac{d\theta(t)}{dt},
    \end{aligned}
    \label{Wand}
\end{equation}
where $\theta(t)$ is a tensor of the NN parameters in time $t$, and $ln(L_{train})$ is the logarithm of the cross-entropy training loss. 
$\eta(t)$ is a noise given by $\langle\eta_i(t)\eta_j(t') \rangle=\lambda (2-\lambda)k_BT\delta_{ij}\delta(t-t')$ , $k_BT=1$ is a temperature term, and $\lambda$ is a friction coefficient.
When sampling the entropy landscape using WLMD, we obtain the corresponding training loss and test accuracy of the NN to locate where to add entropy. 
For the WanD optimizer, its training process samples the entropy landscape $S(ln(L_{train}))$ based solely on training loss, due to the fundamental constraint that it can only access training data.
In addition, we set up an appropriate entropy bias to accelerate the NN's exploration at low training loss area. 
The overall WanD training procedure is summarized in Algorithm \ref{alg:WanD}. 

\begin{algorithm}
\caption{WanD Optimizer}\label{alg:WanD}
\begin{algorithmic}
\Require {Transformer model $T(\cdot )$, the entropy bias $S_{bias}$, training data $(X_{train},y_{train})$and test data$({X_{test},y_{test}})$}
\Ensure {Transformer model $T$ reach generalization: $Accuracy(T(X_{test}),y_{test})\approx100\%$}
\State {The initial entropy: $S \gets S_{bias}$}
\For {each optimize epoch}
\State {Compute the logarithm of the train loss:} 
\State {$ln(L_{train})=ln(CrossEntropyLoss(T(X_{train}),y_{train}))$;}
\State {Compute entropy gradient $x_{grad} = \frac{d S(x_{WL})}{d x_{WL}}$}
\State {Compute $\frac{d^2c_i}{dt^2}=-\frac{\partial ln(L_{train})}{\partial c_i}x_{grad}$}
\For {each parameter $c_i$}
    \State{Update the parameter velocity: $\frac{dc_i}{dt} \gets \frac{dc_i}{dt}(1-c_f)+\frac{d^2c_i}{dt^2}\Delta t+v_{Langevin}$ }
    \State{Update the parameter: $c_i \gets c_i+\frac{dc_i}{dt}\Delta t$}
    
\EndFor
\State{Update the Wang-Landau factor by Equation \ref{entropy update}:  $f_{WL}\gets F_{WL} min\{t/t_{WL},t_{WL}/t\}$ }
\State{Update the entropy landscape by Equation \ref{entropy update}:  $S(x_{WL})\gets S(x_{WL})+f_{WL}\delta (x-x_{WL})$ }
\State {Verification:$A_{test}=Accuracy(T(X_{test}),y_{test})$}
\EndFor
\end{algorithmic}
\end{algorithm}

\begin{figure}[htb]
  \centering
  \includegraphics[width=\linewidth]{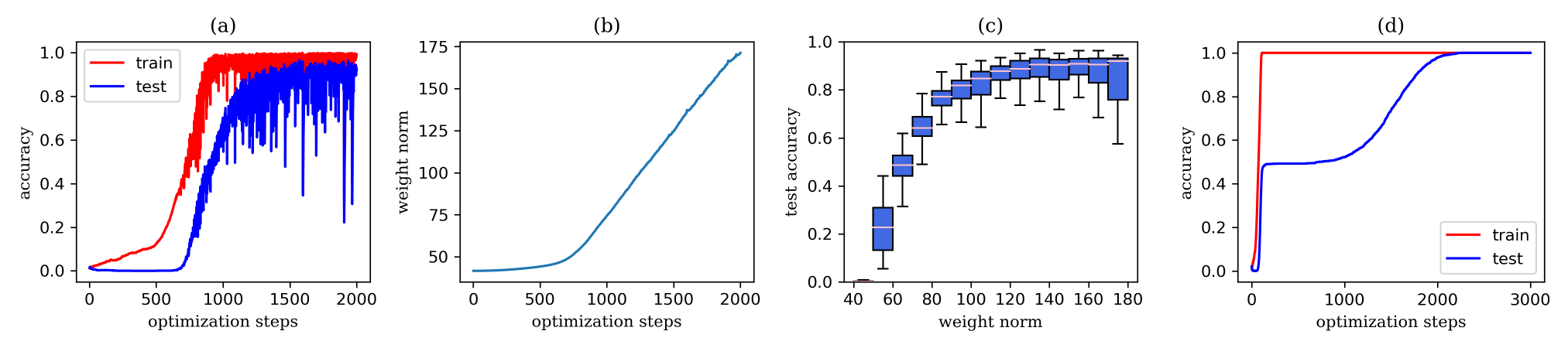}
  \caption{Using (a)(b)(c) WanD and (d) AdamW to train a one layer Transformer on modular addition ($p=67$) task. (a) Train and test accuracy over time. The grokking is largely eliminated while training with WanD. (b) Weight norm over time. (c) Test accuracy over weight norm. These (a, b, c) together demonstrate that the model can generalize at high weight norm. (d) Train and test accuracy over time. Both WanD and AdamW use the same learning rate. }
  \label{fig:WandandAdamW}
\end{figure}

The training results for modular addition task using WanD or AdamW are shown in Figure \ref{fig:WandandAdamW}. 
Both methods have comparable performance, achieving generalization after $\sim 2,000$ optimization steps. 
WanD has a shorter running time, taking about 50 seconds per 1K steps compared to roughly 72 seconds for AdamW on a single GeForce RTX 4090 GPU.
During the training process of WanD, the weight norm of the NN parameters continued to increase, eventually reaching about 175, indicating that the NN can generalize at high weight norm.
This agrees with the work of Kumar et al. \cite{kumar2023grokking} that regularization is not necessary for grokking, but in a more strictly-defined context as we do not scale NN’s output, which could be taken as effectively changing the weight norm.
We present additional training results for WanD and AdamW in the Appendix \ref{dif_split_fraction} on the same task but eliminating grokking by increasing the training dataset split fraction.

\section{Conclusions and Discussion}
\label{discussion and conclusion}

\subsection{Conclusions}

In this work, we investigated grokking in the modular arithmetic tasks using a one-layer transformer, and found that Boltzmann entropy (logarithm  of solution volume) as an excellent indicator of generalization.
By sampling the entropy landscape of the network on training loss and test accuracy, we found that i). there is no entropy barrier between the memorization state and the generalization state, which means that the memorization state is not in an equilibrium or meta-stable state. 
Why does grokking take so long to happen even without crossing free energy barriers? 
In the last part of this section, we postulate that the reason is the formation of glass states.
ii). The equilibrium test accuracy, as a function of the training loss, appears continuous. This strongly suggests that that grokking is not a first-order phase transition as argued by previous work \cite{rubin2023grokking}, because a first-order phase transition implies discontinuity of physical quantities.
Our results also indicate that the equilibrium state in the entropy landscape has a significant advantage over the training curve, which we refer to it as the high-entropy advantage under grokking. 
Such a high-entropy advantage is weakened but still apparent when grokking is eliminated via fixing weight norm. This demonstrates a correlation between equilibrium state and NNs' generalization, consistent with the conclusion of Yang et al.\cite{yang2025high}. 
In addition, we designed a toy optimizer based on the Wang-Landau molecular dynamics method, WanD, which encourages exploration of the high entropy state for NNs' training. 
WanD successfully achieves generalization with a similar time efficiency as AdamW but avoids grokking. 
The generalization states found by WanD have a much higher weight norm compared with the one needed to eliminate grokking in traditional training \cite{liu2022omnigrok}.
This further supports the conclusions of Kumar et al. \cite{kumar2023grokking} that grokking is not solely due to weight decay as some theory proposed before \cite{liu2022omnigrok} in a strictly-defined context. 

\subsection{Discussion}
Based on the above conclusions, we proposed a hypothesis: grokking is similar to the process of quickly cooling a liquid to form a glass and then slowly letting it relax back to the equilibrium state. 
Through this analogy, we try to answer two key questions about grokking:

\textbf{What causes the NN to fall into the memorization state?}
One requirement for the formation of a glass is to cool the liquid below the glass transition temperature \cite{hansen2013theory}. 
Below this temperature, the free movement of atoms is greatly restricted, thus hindering the system from evolving towards equilibrium.
In grokking tasks such as modular arithmetics, the NNs are severely overparameterized, i.e., the number of the parameters is much larger than the size of the dataset.
During training, the NN can fully memorize all training data and achieve a very low training loss.
If we take the training loss as the energy, this is equivalent to lowering the temperature below the glass transition temperature.
Another requirement for glass formation is a sufficiently fast cooling rate, so that the molecules do not have time to rearrange themselves into the lower-free-energy crystalline structure.
During the training process, the gradient descent based optimizers, especially adaptive optimizer like Adam, effectively implement a rapid 'cooling' process  by aggressively driving parameters towards regions of reduced training loss.
This rapid descent often proceeds with insufficient perturbations (caused by large learning rate, small batch size, and data accuracy, etc.) to escape from poor local minima, making NNs easier to reside in a non-generalizable loss minima and therefore form a computational glass. 
In contrast, the Langevin dynamics in the WanD optimizer we designed enriches the perturbations in training, thus avoiding the formation of computational glass.

\textbf{What is the internal dynamics of the grokking process?}
When a liquid is rapidly cooled, it can form a non-equilibrium glassy state that gradually approaches equilibrium through structural relaxation—a slow kinetic process \cite{riechers2022predicting}. 
Due to the sluggish motion of atoms at low temperatures, the relaxation time can far exceed the timescale of the cooling process. 
Analogously, in grokking tasks, the equilibrium state at low training loss corresponds to good generalization. 
However, parameters located in the low train loss regime exhibit very small gradients, which severely limits the ability of gradient-based optimizers to effectively update the network toward this generalizing equilibrium. 
This dynamical slowdown results in a substantial delay between the memorization and generalization phases. 
The effect is further exacerbated by adaptive optimizers like Adam, where the non-uniform scaling of gradients can further dampen parameter updates, increasing the likelihood of being trapped in non-generalizable minima \cite{zhou2020towards}. 
Although the duration of the grokking process varies across training runs, as observed in prior studies and our experiments, the transition to generalization consistently takes significantly longer than the memorization stage \cite{power2022grokking}, mirroring the slow relaxation behavior characteristic of glassy systems \cite{riechers2022predicting}.

Our conclusions agree and disagree with some previous studies. 
There were a series of work arguing that grokking is due to a transition from poor to good representation learning \cite{kumar2023grokking, varma2023explaining, kunin2024get, liu2022towards}.
This agrees with our analogy of relaxation from a poorly structured ``glassy'' (memorizing) state to a more ordered (generalizing) configuration.
In fact, our entropy perspective offers a resolution to debates regarding solution ``efficiency'' and parameter norms in previous works \cite{kumar2023grokking, varma2023explaining}.
While Varma et al.\cite{varma2023explaining} argue that grokking requires a transition to an ``efficient'' generalizing circuit characterized by a lower parameter norm, Kumar et al.\cite{kumar2023grokking} challenges it as they identified generalizing solutions with higher norm than memorizing solutions. 
Our framework suggests this conflict stems from using parameter norm alone to define efficiency, while entropy is actually a better indicator since the efficient solution will have more ``free'' parameters and therefore a larger entropy.
This is also consistent with Merrill et al. \cite{merrill2023tale} who showed that memorization solution and generalization solution correspond to dense and sparse subnetworks, respectively.
In the sparse subnetworks, only a few neurons participate in the memorization of the rules, and the remaining degrees of freedom contribute to the entropy, so it should have a larger entropy as we observed. 
Kumar et al. \cite{kumar2023grokking} proposed that the key determinant of grokking is the rate of feature learning, which is consistent with our glass analogy, where the rate-limited free motion of particles is an important reason for the glass formation and long relaxation time.
The finding by Liu et al. \cite{liu2022towards} that a fast-learning decoder can help trigger grokking also supports our analogy, as the rapid decoder acts similarly to ‘fast cooling’, trapping the network in a glassy memorization state.
Our work also supports the finding that perturbation is beneficial to the grokking learning process \cite{thilak2022slingshot}, as WanD is able to eliminate grokking directly.
Rubin et al. \cite{rubin2023grokking} claimed that grokking is a first-order phase transition process, but our results overturn this conclusion.
This is probably because their network architecture and training data are different from ours, while ours are consistent with the original paper that discovered grokking.
In addition, the high-weight norm generalization states we found through WanD also challenge the conclusions of Liu et al. \cite{liu2022omnigrok}.
Due to space limitations, we provide additional discussion in the Appendix \ref{additional discussion}.

\subsection{Limitation and future work}
\textbf{Limitation}: i). \textbf{Task scope}: Our study is limited to one-layer transformers on modular arithmetic tasks and does not include other grokking benchmarks (e.g., tasks in \cite{rubin2023grokking, liu2022omnigrok}).
ii). \textbf{Interaction with explicit regularization}: The relationship between ``glassy dynamics'' and various explicit regularization methods in grokking systems remains unclear and requires further exploration.
iii). \textbf{Proof-of-concept nature of WanD optimizer}: WanD has only been tested in arithmetic task scenarios, is sensitive to hyperparameters, and exhibits instability near generalization.

Our hypothesis of grokking and glass relaxation invites further investigation. 
While we focused on experiments with the original grokking tasks \cite{power2022grokking}, future studies should examine whether our findings are applicable for other grokking tasks.
Furthermore, a recent study has compared the behavior of deep neural network and structural glass with quantitative measurements \cite{winter2025glassy}, and we can refer to its research method to verify whether the grokking NN is a structural glass. 
In addition, it remains unclear how does glassy dynamics affect the effectiveness of regularization methods in removing grokking. 
Finally,  we also aim to further improve the WanD optimizer to enhance its stability and general applicability. 
As an algorithm based on WLMD, WanD involves multiple hyperparameters that require tuning. 
Our experiments identify the learning rate, temperature, and friction coefficient as the three most influential factors affecting training outcomes.
We plan to further investigate empirical guidelines for effective hyperparameter selection.
Techniques from statistical physics designed to accelerate equilibration, such as simulated annealing and parallel tempering, can also be applied to WanD's future refinement \cite{frenkel2023understanding}. 
We anticipate that approaches such as a dynamic hybrid optimization strategy combining WanD and AdamW could help mitigate instability issues and present some preliminary results in the Appendix \ref{dif_split_fraction}.
Ultimately, we aim to develop WanD into a general-purpose optimizer applicable to a broader range of machine learning tasks.
\clearpage

\subsection*{Acknowledgments}
The authors thank National Natural Science Foundation of China for supporting this
research (Grant 12405043, G.Z.).We also thank computational resources provided by Bridges-2 at
Pittsburgh Supercomputing Center through ACCESS allocation CIS230096 (E.Y.).
\bibliographystyle{unsrt}
\clearpage
\appendix

\section{Additional Experimental Details}

\label{additional detials}
\subsection{Model and Datasets}
\label{model_and_datasets}

We followed the work of Nanda et al.\cite{nanda2023progress} for the model setup of modular arithmetic tasks. 
We use the same one-layer transformer model for the programs of AdamW optimizer training, WanD optimizer training, and WLMD entropy sampling. 
This transformer model has 4 attention heads by default, each with a width of 32, yielding a model dimension of 128, and attached with an MLP layer with a width of 512. 
In the experiment of eliminating grokking by controlling the weight norm, we modified the attention head width (8/16/32) and the MLP layer width (128/256/512). 
In order to reduce the computational cost of WLMD entropy sampling, we selected a smaller prime number 67 as the modulus for our task. 
Although some works \cite{liu2022omnigrok, nanda2023progress} choose 113 as the modulus, there is still exist grokking in arithmetic tasks with 67 as the modulus, so we think this setting is appropriate.
We also use the cross-entropy loss as our loss function. 
For more details on the model’s hyperparameters, please refer to the provided code.

\subsection{Training Details and Cost}
\label{training_details_and_cost}

For modular arithmetic tasks training by AdamW, we use a full batch size on 1 GeForce RTX 4090 GPU, training for 10000 optimization steps with weight decay 1 or fixed weight norm 30.
The hyperparameters for training with AdamW are given in Table \ref{tab:AdamW training hypeparameters}.
For our toy optimizer WanD training, since it is not yet fully developed, there are many hyperparameters that directly affect the generalization of the training results. 
Among them, the learning rate, temperature, and friction coefficient are the three key factors that determine the training results.
We provide a set of hyperparameter combinations in the code that can generalized well.

\begin{table}[h!]
  \caption{AdamW training hyperparameters}
  \label{tab:AdamW training hypeparameters}
  \centering
  \setlength{\tabcolsep}{4pt}
  \begin{tabular}{lccccl}
    \toprule
    Operation & Model Dimension & MLP Width & Learning Rate & Weight Decay & Weight Norm\\
    \cmidrule(r){1-6}
    $x+y$       & 32    & 128   & 1e-3  & 0 & Fixed at 30   \\  
    $x+y$       & 64    & 256   & 1e-3  & 0 & Fixed at 30   \\ 
    $x+y$       & 128   & 512   & 1e-3  & 0 & Fixed at 30   \\ 
    $x+y$       & 128   & 512   & 3e-3  & 1 & Not fixed     \\ 
    $x^2+y$     & 128   & 512   & 5e-3  & 1 & Not fixed     \\ 
    $x^3+xy^2+y$& 128   & 512   & 1e-2  & 1 & Not fixed     \\ 
    \bottomrule
  \end{tabular}
\end{table}

For WLMD sampling, we use 4 GPUs to run 8 processes. 
Each process uses the same data split and updates the entropy landscape according to a different random seed. 
A single process runs 100,000 epochs at a time, then calculates the average entropy of all processes and updates the entropy landscape. 
When running 8 WLMD processes on 4 GeForce RTX 4090 GPUs, it takes about a day for each process to update 1M epochs. 
When running 8 WLMD processes on 4 NVIDIA H100 GPUs, it takes about 16 to 17 hours for each process to update 1M epochs. 
In order for the entropy landscape to converge to its true distribution, tens to hundreds of millions of epochs are typically required. 
For detailed hyperparameter settings, please refer to the code we provide.

\subsection{WLMD details}
\label{WLMD details}

\textbf{Smoothed accuracy}: 
To ensure well-defined gradient of test accuracy with respect to NN parameters in WLMD, we compute a smoothed test accuracy using a sigmoid function, following \cite{yang2025high}. 
The detailed procedure for obtaining smoothed accuracy is outlined in Algorithm \ref{alg:Smoothed Accuracy}. In our work, we choose the smoothed accuracy slope $s=7$.

\begin{algorithm}
\caption{Smoothed Accuracy}\label{alg:Smoothed Accuracy}
\begin{algorithmic}
\Require {Transformer model $T(\cdot)$, dataset $(X=\{x_1,x_2,...,x_n\},\hat Y=\{\hat y_1,\hat y_2,...,\hat y_n\})$, smoothed accuracy slope $s$}
\Ensure {The value of smoothed accuracy close to the true accuracy, and the gradient of smoothed accuracy with respect to parameters has a finite value.}
\For {each $(x_i,\hat y_i)$ in $(X,\hat Y)$}
\State {Get the highest logit among all wrong output classes: $p_i(y=y_{i\_wrong})$ where $y_{i\_wrong}\in Y_i=T(x_i)$}
\State {Get the logit of the correct output class: $p_i(y=\hat y_i)$}
\State {Compute the smoothed accuracy:$A_i=Sigmoid((p_i(y=\hat y_i)-p_i(y=y_{i\_wrong}))\cdot s)$}
\EndFor
\State {Compute the total smoothed accuracy:$A=\sum{A_i}/{n}$}
\end{algorithmic}
\end{algorithm}

\textbf{Entropy bias}: 
When sampling the entropy landscape of grokking NNs, we are interested in the regime with lower training loss, which corresponds to the memorization state and the generalization state. 
However, areas with high training loss usually have higher entropy (as evidenced by randomly initialized networks that generally possess high training loss), which will waste a lot of time exploring in the early stages of sampling.
In order to prevent the simulation program from wasting time outside the range of our interest, we set an additional entropy bias:
\begin{equation}
    \begin{aligned}
     \mathrm{if}\ ln(L_{train})>L_0,\ S_{bias}(ln(L_{train}))=\beta (ln(L_{train})-L_0)^2;\  \mathrm{else}\ S_{bias}(ln(L_{train}))=0.
    \end{aligned}
    \label{entropy bias}
\end{equation}
If the training loss is outside the range of interest $(ln(L_{train})>L_0)$ when updating the NN parameters, this entropy bias will provide additional entropy gradients to to force the NN to move towards lower training loss. 

The overall WLMD entropy sampling procedure is outlined in Algorithm \ref{alg:WLMD}.
We employ the cross-entropy loss for our modular arithmetic task and a logarithmic scale of the training loss is used to better examine the entropy landscape near zero training loss.

\begin{algorithm}
\caption{WLMD Entropy Sampling Method}\label{alg:WLMD}
\begin{algorithmic}
\Require {Transformer model $T(\cdot )$, its current entropy landscape $S$, the entropy bias $S_{bias}$, training data $(X_{train},y_{train})$and test data$({X_{test},y_{test}})$;}
\Ensure {Convergent entropy landscape $S(x_{WL},y_{WL})$ where $x_{WL} = ln(L_{train})$, $y_{WL} = A_{test}$;}
\State {Initialize: $S \gets S_{bias}$;}
\For {each sampling epoch}
\State {Compute the logarithm of the train loss and smoothed test accuracy:} 
\State {$ln(L_{train})=ln(CrossEntropyLoss(T(X_{train}),y_{train}))$;}
\State {$A_{test}=SmoothedAccuracy(T(X_{test}),y_{test})$;}
\State {Compute entropy gradient $x_{grad} = \frac{\partial S(x_{WL},y_{WL})}{\partial x_{WL}}$ and $y_{grad} =\frac{\partial S(x_{WL},y_{WL})}{\partial y_{WL}}$;}
\State {The combined loss $L_{combined} = -x_{WL}x_{grad}-y_{WL}y_{grad}$;}
\State {Do $L_{combined}.backward()$ and get $\frac{d^2c_i}{dt^2}$;}
\For {each parameter $c_i$}
    \State{Update the parameter velocity by Equation \ref{Langevin motion}: $\frac{dc_i}{dt} \gets \frac{dc_i}{dt}(1-c_f)+\frac{d^2c_i}{dt^2}\Delta t+v_{Langevin}$;}
    \State{Update the parameter: $c_i \gets c_i+\frac{dc_i}{dt}\Delta t$;}
    
\EndFor
\State{Update the Wang-Landau factor by Equation \ref{entropy update}:  $f_{WL}\gets F_{WL} min\{t/t_{WL},t_{WL}/t\}$; }
\State{Update the entropy landscape by Equation \ref{entropy update}:  $S(x_{WL},y_{WL})\gets S(x_{WL},y_{WL})+f_{WL}\delta (x-x_{WL},y-y_{WL})$; }
\EndFor
\State {$S \gets S-S_{bias}$;}
\end{algorithmic}
\end{algorithm}

\section{Additional Results}
\label{additional results}

\subsection{Visualization of modular arithmetic tasks}
\label{visualization}

\begin{figure}[htb]
  \centering
  \includegraphics[width=\linewidth]{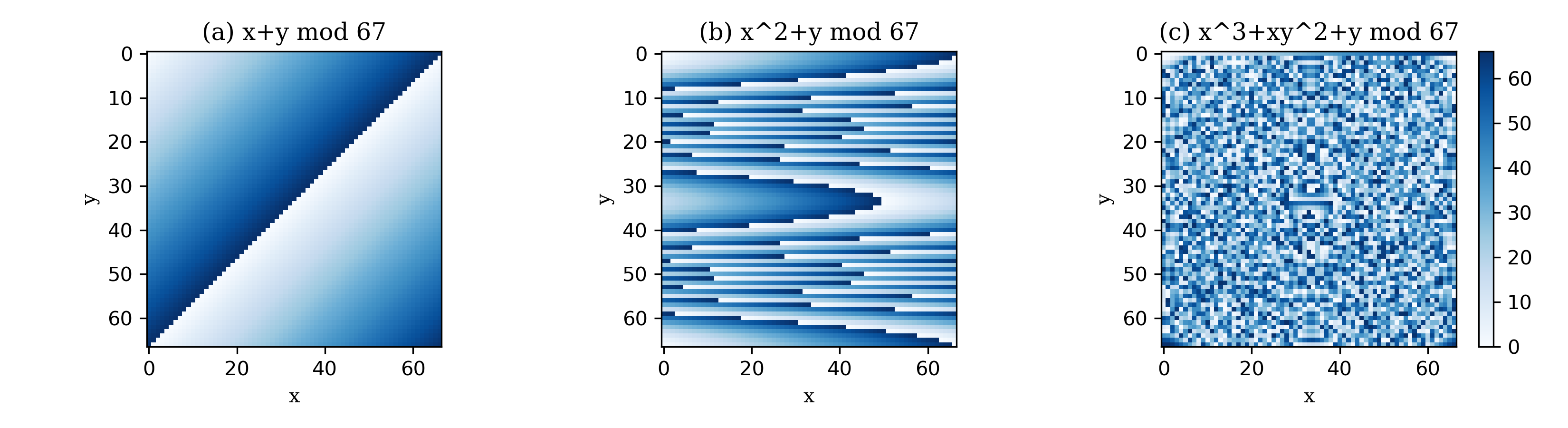}
  \caption{Visualization of three different arithmetic tasks. (a) $x+y\ mod\ 67$. (b) $x^2+y\ mod\ 67$. (c) $x^3+xy^2+y\ mod\ 67$. Our protocol is equivalent to randomly taking 50\% of the pixels from the image as the training set and trying to complete the rest. The more complex the expression, the more visually irregular the image is. }
  \label{fig:visual}
\end{figure}

First, we show a visualization of the three modular arithmetic tasks in the Figure \ref{fig:visual}. 
A complete dataset with 67*67=4489 data points can be viewed as a 67*67 pixel image, where the x-axis and y-axis coordinates of the pixels correspond to the two numbers $\langle x\rangle$ and $\langle y\rangle$ involved in the operation, and the color of the pixel corresponds to the answer $\langle x\circ y\rangle$. 
Half of the pixels are randomly selected as the training set, and the other half as the test set. 
Training the transformer model to learn the modular arithmetic operation is equivalent to completing the entire image according to the regularity of the image after being masked.
In particular, the numbers in the arithmetic operations are just equivalent to different characters, and they can be replaced by any other different characters without affecting the validity of the modular arithmetic task. 
In order to observe the image intuitively, we dye the pixels with light to dark blue according to the corresponding numbers from small to large. 
It is not difficult to observe that the visualization images of simpler tasks have more obvious regularities. 
For the task that cannot be learned, its visualization image is close to random noise. 

\subsection{AdamW training results of modular arithmetic tasks}
\label{AdamW_training_results}

\begin{figure}[htb]
  \centering
  \includegraphics[width=\linewidth]{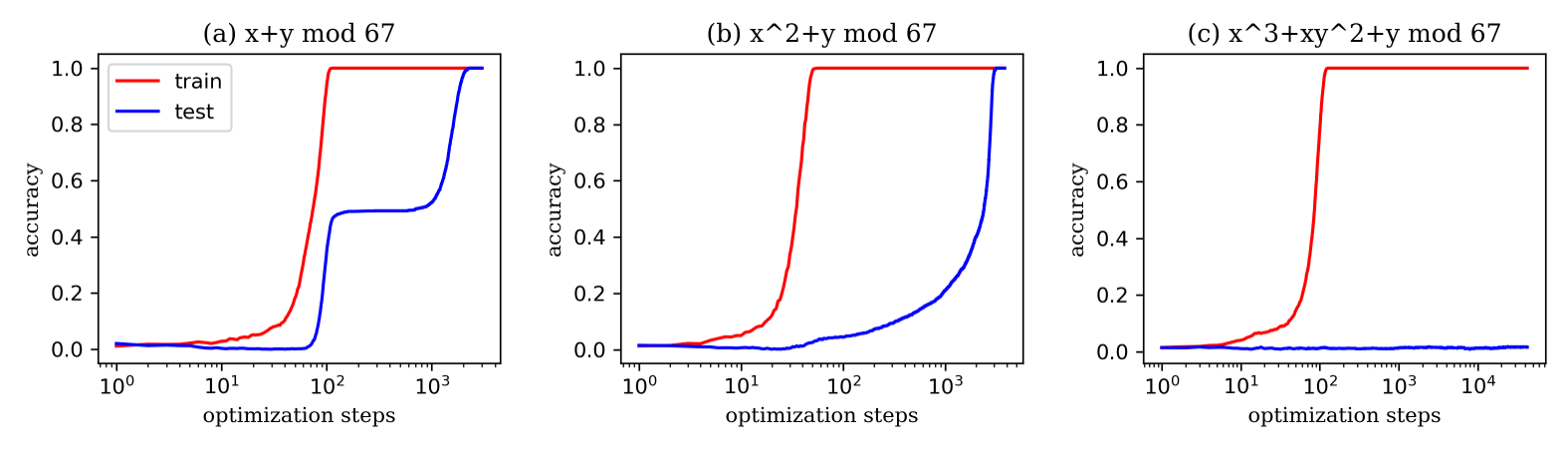}
  \caption{Training results of three different arithmetic tasks: (a) $x+y\ mod\ 67$ is easy to learn, but the hysteresis of the test accuracy shows grokking. (b) $x^2+y\ mod\ 67$ is hard to learn, the grokking is more obvious. (c) $x^3+xy^2+y\ mod\ 67$ cannot be learned. }
  \label{fig:three_tasks_training}
\end{figure}

The AdamW training results of these three modular arithmetic tasks are shown in Figure \ref{fig:three_tasks_training}. 
The figure shows that as the expression becomes more and more complex, the learning difficulty of the NN also increases, from easy to learn, to difficult to learn, to impossible to learn. 
The generalization delay is more significant in the difficult task, which means grokking is more pronounced. 

\begin{figure}[htb]
  \centering
  \includegraphics[width=\linewidth]{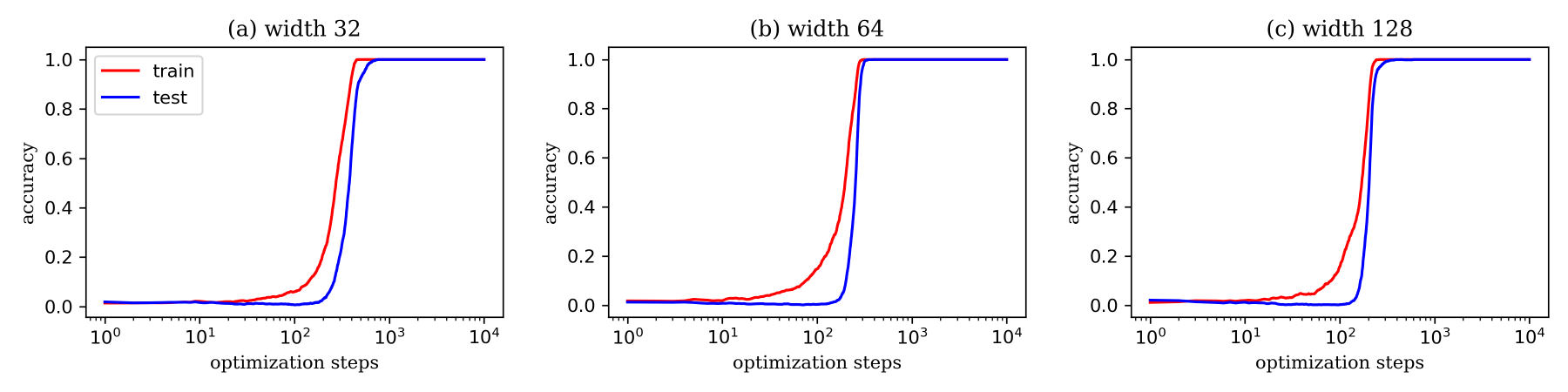}
  \caption{Training results of modular addition ($p=67$) task with three transformer models that have different widths. Here weight norm is constrained to a fixed value of 30. }
  \label{fig:AdamW_training_result}
\end{figure}

We also train the transformer in the modular addition task with a constrained weight norm. 
The weight norm is restricted to the Goldilocks zone \cite{liu2022omnigrok}, and we set it to 30 for three NNs with different widths. 
We tested three NNs with different widths mainly because Yang et al. \cite{yang2025high} observed that the high-entropy advantage becomes more significant with smaller width. 
The training results are shown in Figure \ref{fig:AdamW_training_result}. 
In fact, these three NNs reach generalization after almost the same number of optimization steps, and none of them exhibits grokking. 
Combined with the entropy landscape in Figure \ref{fig:entropywidths} in the main text, we believe that the relationship between high-entropy advantage and NN width is not obvious in the grokking task. 

\subsection{Training results of AdamW and WanD under different dataset splits}
\label{dif_split_fraction}

\begin{figure}[htb]
  \centering
  \includegraphics[width=\linewidth]{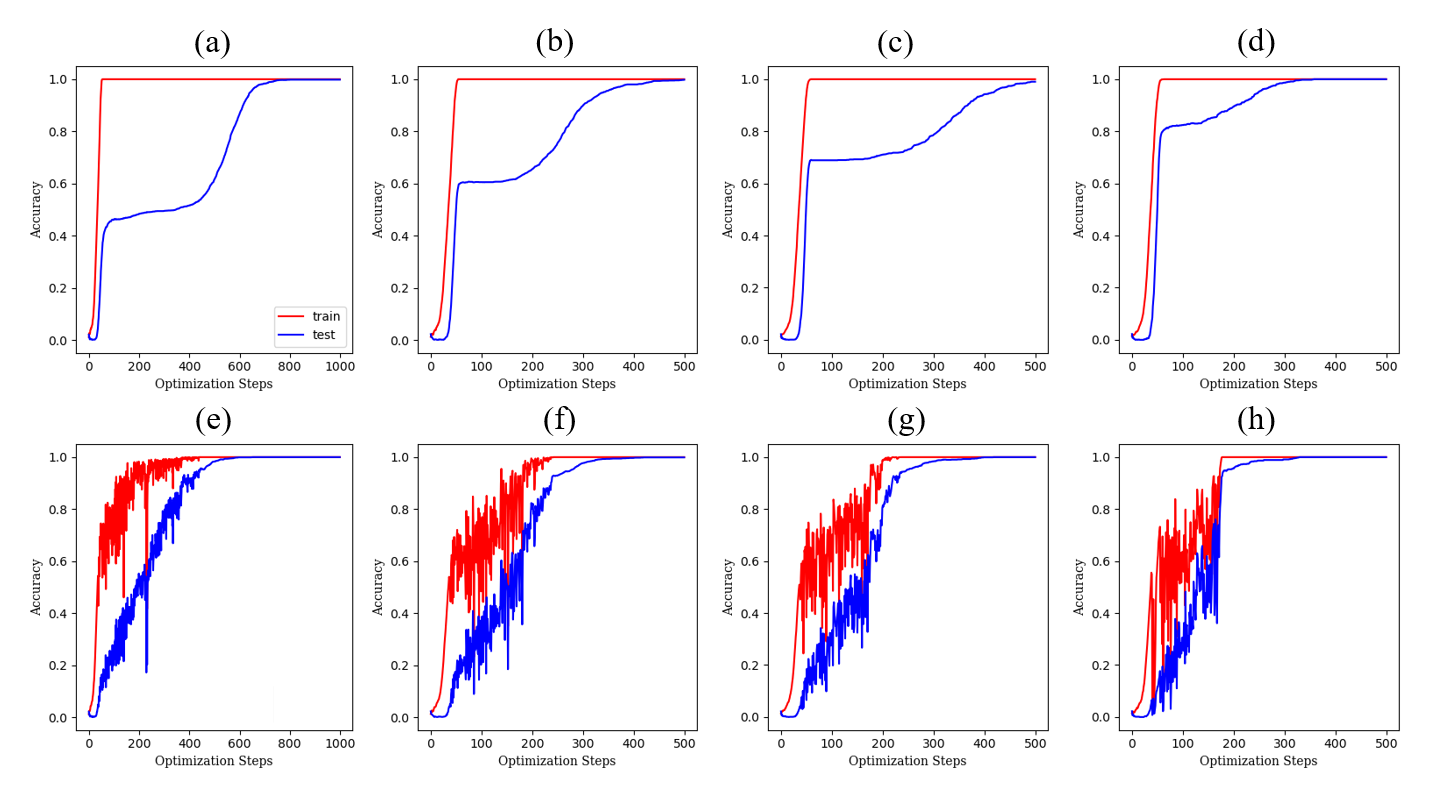}
  \caption{AdamW and WanD training results of $x+y\ mod\ 67$ task with different training dataset fraction. Top row:AdamW; bottom row:WanD; (a)(e) 50\%, (b)(f) 60\%, (c)(g) 70\%, and (d)(h) 80\% training dataset split fraction. The hyperparameter of WanD: kT=0.5, frictionCoefficient=0.5, maxTrainLossToStudy=-8.}
  \label{fig:dif_split_easy}
\end{figure}

\begin{figure}[htb]
  \centering
  \includegraphics[width=\linewidth]{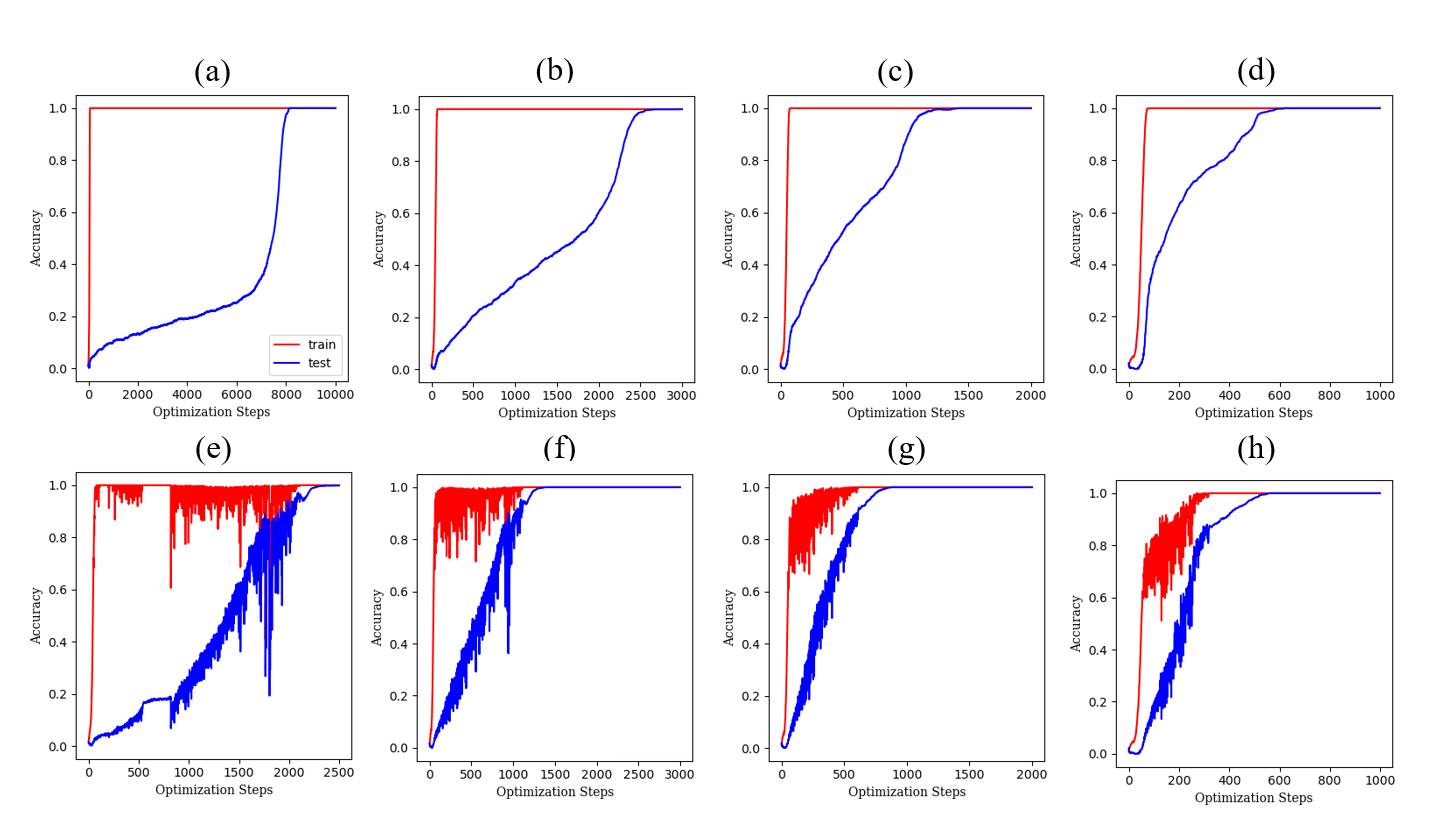}
  \caption{AdamW and WanD training results of $x^2+y\ mod\ 67$ task with different training dataset fraction. (a)(b)(c)(d):AdamW; (e)(f)(g)(h):WanD; (a)(e)/(b)(f)/(c)(g)/(d)(h): 50\%/60\%/70\%/80\% training dataset split fraction. The hyperparameter of WanD: kT=0.7, frictionCoefficient=0.6, maxTrainLossToStudy=-8. It is worth noting that WanD in (e) achieves generalization with nearly three times the efficiency of AdamW in (a). }
  \label{fig:dif_split_hard}
\end{figure}

Increasing the training dataset split fraction is an effective way to eliminate grokking \cite{power2022grokking}. We tested the training results of AdamW and WanD on the $x+y\ mod\ 67$ (Figure \ref{fig:dif_split_easy}) and $x^2+y\ mod\ 67$ (Figure \ref{fig:dif_split_hard}) tasks at the same learning rate 0.003, with the training dataset fraction being 50\%, 60\%, 70\%, and 80\%, respectively.
Here we anneal WanD with AdamW so that the update weight of WanD increases first and then decreases during training, thus avoiding the instability problem at the end of WanD training.
Specifically, in each iteration, the NN parameters are updated by the WanD optimizer with weight $\alpha$ and by the AdamW optimizer with weight $1-\alpha$, where $\alpha=(2A_{test}-1)^2$.

The results show that WanD has a significant training efficiency advantage in tasks with severe grokking, and even in tasks where grokking is greatly eliminated, WanD can still achieve generalization ahead of AdamW.
This result demonstrates the great potential of WanD as a toy optimizer and motivates us to continue improving it.

\section{Additional Discussion}
\label{additional discussion}

\textbf{Robustness of entropy landscape diagrams:}
In this work, we visualize the entropy landscape with respect to the logarithm of the training cross-entropy loss and the smoothed test accuracy. 
Due to the symmetry between training and test datasets, modifying the training loss formulation is conceptually equivalent to altering the test loss.
Our reported results are based on the smoothed test accuracy, whose inverse can be interpreted as a form of loss function. 
We have also conducted internal experiments using the cross-entropy test loss in place of the smoothed accuracy. 
The resulting entropy diagram consistently support the same conclusion: entropy barrier is not observed, and only a single connected high-entropy region exists, indicating no sign of a first order phase transition. 
This consistency confirms that our findings are robust to the specific choice of the loss function.

\textbf{Sparse subnetworks:} As shown in \cite{merrill2023tale}, grokking is associated with sparse subnetworks in which only a subset of neurons is involved in rule memorization. 
From an entropy perspective, parameters outside these functional subnetworks do not affect generalization.
Perturbing such parameters preserves model performance, implying that the corresponding solution set occupies a larger volume in parameter space. 
Since entropy is defined as the logarithm of this volume, neural networks with sparse subnetworks exhibit higher entropy.

\textbf{Other entropy-based optimizer:} WanD is an optimizer built on the one-dimensional WLMD framework.
It excels at identifying high-entropy states, particularly the generalized solution, via molecular dynamics, leveraging the fact that generalized solutions possess higher entropy at low training loss. 
Prior to our work, there is another entropy-guided optimizers named Entropy-SGD \cite{chaudhari2019entropy} have been proposed. Below, we briefly highlight the theoretical distinctions between WanD and Entropy-SGD:

\begin{itemize}
\item \textbf{Different notions of entropy.}
The core distinction lies in the entropy definition: WanD targets the global Boltzmann entropy, reflecting the total volume of the solution space, whereas Entropy-SGD focuses on local entropy, which captures the flatness of individual minima.
\item \textbf{Different methodologies.}
WanD estimates entropy via 1D WLMD sampling, while Entropy-SGD relies on Langevin dynamics to approximate local entropy gradients.
\end{itemize}
Optimizers like Adam often progress slowly due to limited gradient signals during the memorization-to-generalization transition \cite{kumar2023grokking}. 
In principle, Entropy-SGD could potentially accelerate grokking if its identified high local-entropy states exhibit strong generalization. 
A rigorous empirical comparison between these approaches remains a valuable direction for future work.


\end{document}